\documentclass[conference]{IEEEtran}
\IEEEoverridecommandlockouts

\usepackage{cite}
\usepackage{amsmath,amssymb,amsfonts}
\usepackage{graphicx}
\usepackage{textcomp}
\def\BibTeX{{\rm B\kern-.05em{\sc i\kern-.025em b}\kern-.08em
    T\kern-.1667em\lower.7ex\hbox{E}\kern-.125emX}}
\usepackage[utf8]{inputenc}
\usepackage{amsmath}
\usepackage{amssymb}
\usepackage{amsfonts}
\usepackage{hyperref}

\usepackage{amsmath}
\usepackage{color}
\usepackage{amssymb,amsmath,latexsym,amsfonts,amsthm}
\usepackage{mathtools}

\usepackage{makecell}

\usepackage{subcaption}
\usepackage{float}
\usepackage{xcolor}
\usepackage{multirow}
\usepackage{booktabs}
\usepackage{mathrsfs}  

\usepackage[linesnumbered,ruled,vlined]{algorithm2e}
\usepackage{float}
\usepackage{upgreek}

\usepackage{multicol}

\usepackage{caption}
\usepackage{subcaption}

\def\BibTeX{{\rm B\kern-.05em{\sc i\kern-.025em b}\kern-.08em
    T\kern-.1667em\lower.7ex\hbox{E}\kern-.125emX}}
\begin{document}

\title{Unreasonable Effectiveness of Last Hidden Layer Activations for Adversarial Robustness}

\author{\IEEEauthorblockN{Omer Faruk Tuna}
\IEEEauthorblockA{\textit{Computer Engineering Department} \\
\textit{Isik University}\\
Istanbul, Turkey \\
omer.tuna@isikun.edu.tr}
\and
\IEEEauthorblockN{Ferhat Ozgur Catak}
\IEEEauthorblockA{\textit{Electrical Engineering and Computer Science} \\
\textit{University of Stavanger}\\
Rogaland, Norway \\
f.ozgur.catak@uis.no}
\and
\IEEEauthorblockN{M. Taner Eskil}
\IEEEauthorblockA{\textit{Computer Engineering Department} \\
\textit{Isik University}\\
Istanbul, Turkey \\
taner.eskil@isikun.edu.tr}
}

\maketitle





\begin{abstract}
In standard Deep Neural Network (DNN) based classifiers, the general convention is to omit the activation function in the last (output) layer and directly apply the softmax function on the logits to get the probability scores of each class. In this type of architectures, the loss value of the classifier against any output class is directly proportional to the difference between the final probability score and the label value of the associated class. Standard White-box adversarial evasion attacks, whether targeted or untargeted, mainly try to exploit the gradient of the model loss function to craft adversarial samples and fool the model. In this study, we show both mathematically and experimentally that using some widely known activation functions in the output layer of the model with high temperature values has the effect of zeroing out the gradients for both targeted and untargeted attack cases, preventing attackers from exploiting the model's loss function to craft adversarial samples. We've experimentally verified the efficacy of our approach on MNIST (Digit), CIFAR10 datasets. Detailed experiments confirmed that our approach substantially improves robustness against gradient-based targeted and untargeted attack threats. And, we showed that the increased non-linearity at the output layer has some additional benefits against some other attack methods like Deepfool attack.
\end{abstract}

\begin{IEEEkeywords}
trustworthy AI, adversarial machine learning, deep neural networks, robustness
\end{IEEEkeywords}

\section{Introduction}

By the end of 2013, researchers found out that DNN models are vulnerable to well-crafted malicious perturbations. Szegedy et al. \cite{szegedy2014intriguing} were the first to recognize the prevalence of adversarial cases in the context of image classification. Researchers have shown that a slight alteration in an image can influence the prediction of a DNN model. It is demonstrated that even the most advanced classifiers can be fooled by a very small and practically undetectable change in input, resulting in inaccurate classification.
Since then, a lot of research studies \cite{tuna2021exploiting,ilyas2019prior,tuna2020closeness,meng2017magnet} were performed in this new discipline known as \textit{Adversarial Machine Learning} and these studies were not limited just to image classification task. To give some example, Sato et al. \cite{sato2018interpretable} showed in the NLP domain that changing just one word from an input sentence can fool a sentiment analyser trained with textual data. Another example is in the audio domain \cite{carlini2018audio},where the authors generated targeted adversarial audio samples in autonomous speech recognition task by introducing very little distortion to the original waveform. The findings of this study indicate that the target model can simply be exploited to transcribe the input as any desired phrase.

Attacks that take advantage of DNN's weakness can substantially compromise the security of these machine learning (ML)-based systems, often with disastrous results. Adversarial evasion attacks mainly work by altering the input samples to increase the likelihood of making wrong predictions. These attacks can cause the model's prediction performance to deteriorate since the model cannot correctly predict the actual output for the input instances. In the context of medical applications, a malicious attack could result in an inaccurate disease diagnosis. As a result, it has the potential to impact the patient's health, as well as the healthcare industry \cite{finlayson2019adversarial}. Similarly, self-driving cars employ ML to navigate traffic without the need for human involvement. A wrong decision for the autonomous vehicle based on an adversarial attack could result in a tragic accident. \cite{sitawarin2018darts,morgulis2019fooling}. Hence, defending against malicious evasion attacks and boosting the robustness of ML models without sacrificing clean accuracy is critical. Presuming that these ML models are to be utilized in crucial areas, we should pay utmost attention to ML models' performance and the security problems of these architectures.

In principle, adversarial strategies in evasion attacks can be classified based on multiple criteria. Based on the attacker's ultimate goal, attacks can be classified as targeted and untargeted attacks. In the former, the attacker perturbs the input image, causing the model to predict a class other than the actual class. Whereas in the latter, the attacker perturbs the input image so that a particular target class is predicted by the model. Attacks can also be grouped based on the level of knowledge that the attacker has. If the attacker has complete knowledge of the model like architecture, weights, hyper-parameters etc., we call this kind of setting as White-Box Settings. However, if the attacker has no information of the deployed model and defense strategy, we call this kind of setting as Black-Box Settings \cite{NEURIPS2018_e7a425c6}. 

This research study focused on both targeted and untargeted attacks in a White-Box setting.  We propose an effective modification to the standard DNN-based classifiers by adding special kind of non-linear activation functions (sigmoid or tanh) to the last layer of the model architecture. We showed that training a model using high temperature value at the output layer activations and using the model by discarding the temperature value at inference time provides a very high degree of robustness to loss-based White-box targeted and untargeted attacks, together with attacks acting like Deepfool. We hereby name our proposed models as \texttt{Squeezed Models}. Our codes are released on GitHub \footnote{\url{https://github.com/author-name/xxx}} for scientific use.

To summarize; our main contributions for this study are:

\begin{itemize}

    \item We propose an effective modification to standard DNN based classifiers, which enables natural robustness to gradient-based White-box targeted and untargeted attacks.
    
	\item We show that using a specific type of non-linear activation functions at the output layer with high temperature values can actually provide robustness to the model without impairing the ability to learn.
	
	\item We experimentally showed that adding non-linearity to the last hidden layer provides robustness to other types of attacks, like Deepfool.
\end{itemize}

\section{Related Works}
\label{ch:related_work}

Since the uncovering of DNN's vulnerability to adversarial attacks \cite{szegedy2014intriguing}, a lot of work has gone into inventing new adversarial attack algorithms and defending against them by utilizing more robust architectures \cite{HUANG2020100270,catak2020generative,9003212,9099439}. We will discuss some of the noteworthy attack and defense studies separately.

\subsection{Adversarial evasion attacks}

\begin{figure}[!htbp]
    \centering
    \includegraphics[width=1.0\linewidth]{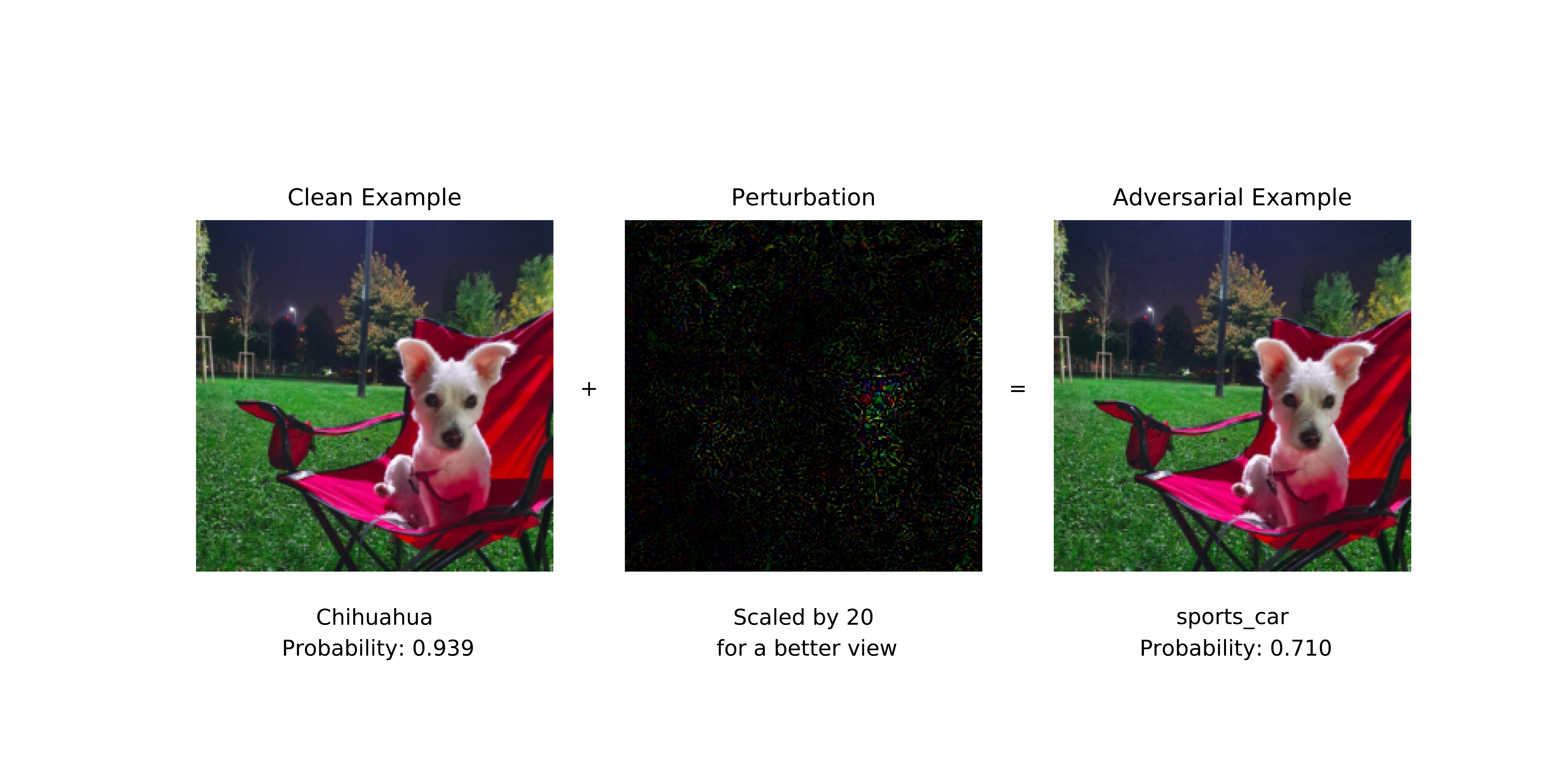}
    \caption{The figure depicts an example to adversarial attack. The original image is subjected to the adversarial perturbation. The precisely crafted perturbation manipulates the model in such a way that a "Dog (Chihuahua)" is wrongly identified as "Sports Car" with high confidence.}
    \label{fig:adv-ml-ex}
\end{figure}

DNN models have some vulnerabilities that make them challenging to defend in adversarial settings. For example, they are mostly sensitive to slight changes in the input data, leading to unexpected results in the model's predictions. Figure \ref{fig:adv-ml-ex} depicts how an adversary could take advantage of such a vulnerability and fool the model using properly crafted perturbation applied to the input.

An important portion of the attack methods are gradient-based and based on perturbing the input sample in order to maximize the model's loss. In recent years, many different adversarial attack techniques have been suggested in literature. The most widely known and used adversarial attacks are \texttt{Fast-Gradient Sign Method, Iterative Gradient Sign Method, \texttt{DeepFool}} and \texttt{Carlini-} \texttt{Wagner}. These adversarial attack algorithms are briefly explained in Section \ref{sec:fgsm-definition} - \ref{sec:carlini}.

\subsubsection{Fast-Gradient Sign Method}\label{sec:fgsm-definition}

This method, also known as FGSM \cite{goodfellow2015explaining}, was one of the first and most well-known adversarial attacks. The derivative of the model's loss function with respect to the input sample is exploited in this attack strategy to determine which direction the pixel values in the input image should be changed in order to minimize the loss function of the model. Once this direction is determined, it changes all pixels in the opposite direction at the same time to maximize loss. One can craft adversarial samples for a model with a classification loss function represented as $J(\theta,\mathbf{x},y)$ by utilizing the formula below, where $\theta$ denotes the parameters of the model, $\mathbf{x}$ is the benign input, and $y_{true}$ is the real label of our input.

\begin{equation}
    \mathbf{x}^{adv} = \mathbf{x} + \epsilon \cdot sign\left(\nabla_x J(\theta,\mathbf{x},y_{true}) \right)
\label{eq:fgsm_untargeted}    
\end{equation}

In \cite{kurakin2017adversarial}, the authors presented a targeted variant of FGSM referred to as the Targeted Gradient Sign Method (TGSM). This way, they could change the attack to try to convert the model's prediction to a particular class. To achieve this, instead of maximizing the loss with respect to the true class label, TGSM attempts to minimize the loss with respect to the target class $J_{target}$. 

\begin{equation}
    \mathbf{x}^{adv} = \mathbf{x} - \epsilon \cdot sign\left(\nabla_x J(\theta,\mathbf{x},y_{target}) \right)
\label{eq:fgsm_targeted}    
\end{equation}

Different from Eq. \ref{eq:fgsm_untargeted}, we now subtract the crafted perturbation from the original image as we try to minimize the loss this time. If we want to increase the efficiency of this approach, we can modify above equation as in Eq.\ref{eq:fgsm_targeted_enhanced}.The only difference is that instead of minimizing the loss of the target label, we maximize the loss of the loss of the true label and also minimize the loss for the alternative label.

\begin{equation}
    \mathbf{x}^{adv} = \mathbf{x} + \epsilon \cdot sign\left(\nabla_x (J(\theta,\mathbf{x},y_{true})-J(\theta,\mathbf{x},y_{target})) \right)
\label{eq:fgsm_targeted_enhanced}    
\end{equation}

\subsubsection{Iterative Gradient Sign Method}
Kurakin et al. \cite{kurakin2017adversarial} proposed a minor but significant enhancement to the FGSM. Instead of taking one large step $\epsilon$ in the direction of the gradient sign, we take numerous smaller steps $\alpha$ and utilize the supplied value $\epsilon$ to clip the output in this method. This method is also known as the Basic Iterative Method (BIM), and it is simply FGSM applied to an input sample iteratively. Equation \ref{eq:bim}  describes how to generate perturbed images under the $l_{inf}$ norm for a BIM attack.

\begin{equation}
\begin{aligned}
\mathbf{x}_{t}^* & = \mathbf{x} \\
\mathbf{x}_{t+1}^* & = clip_{x, \epsilon} \{ \mathbf{x}_{t}  + \alpha \cdot sign \left( \nabla_\mathbf{x} J(\theta, \mathbf{x}_t^*, y_{true}) \right) \}
\end{aligned}
\label{eq:bim}
\end{equation}
where $\mathbf{x}$ is the clean sample input to the model, $\mathbf{x}^*$ is the output adversarial sample at $i$\textsuperscript{th} iteration, $J$ is the loss function of the model, $\theta$ denotes model parameters, $y_{true}$ is the true label for the input, $\epsilon$ is a configurable parameter that limits maximum perturbation amount in given $l_{inf}$ norm, and $\alpha$ is the step size.

As in the case of TGSM, we can easily modify Eq. \ref{eq:bim} to produce targeted variant of BIM. At each intermediate step, we can try to minimize the loss with respect to target class while at the same time maximizing the loss with respect to original class as in Eq. \ref{eq:bim_targeted}.

\begin{equation}
\begin{aligned}
\mathbf{x}_{t}^* & = \mathbf{x} \\
arxiv \mathbf{x}_{t+1}^* & = clip_{x, \epsilon} \{ \mathbf{x}_{t}  + \alpha \cdot sign ( \nabla_\mathbf{x} (J(\theta, \mathbf{x}_t^*, y_{true})-\\
& J(\theta, \mathbf{x}_t^*, y_{target})) ) \}
\end{aligned}
\label{eq:bim_targeted}    
\end{equation}

\subsubsection{Deepfool Attack} \label{sec:deepfool-definition}

This attack method has been introduced by Moosavi-Dezfooli et al. \cite{moosavidezfooli2016deepfool} and it is one of the most strong untargeted attack algorithms in literature. It's made to work with several distance norm metrics, including $l_{inf}$ and $l_{2}$ norms.

\begin{figure}[h!]
    \centering
    \includegraphics[width=0.75\linewidth] {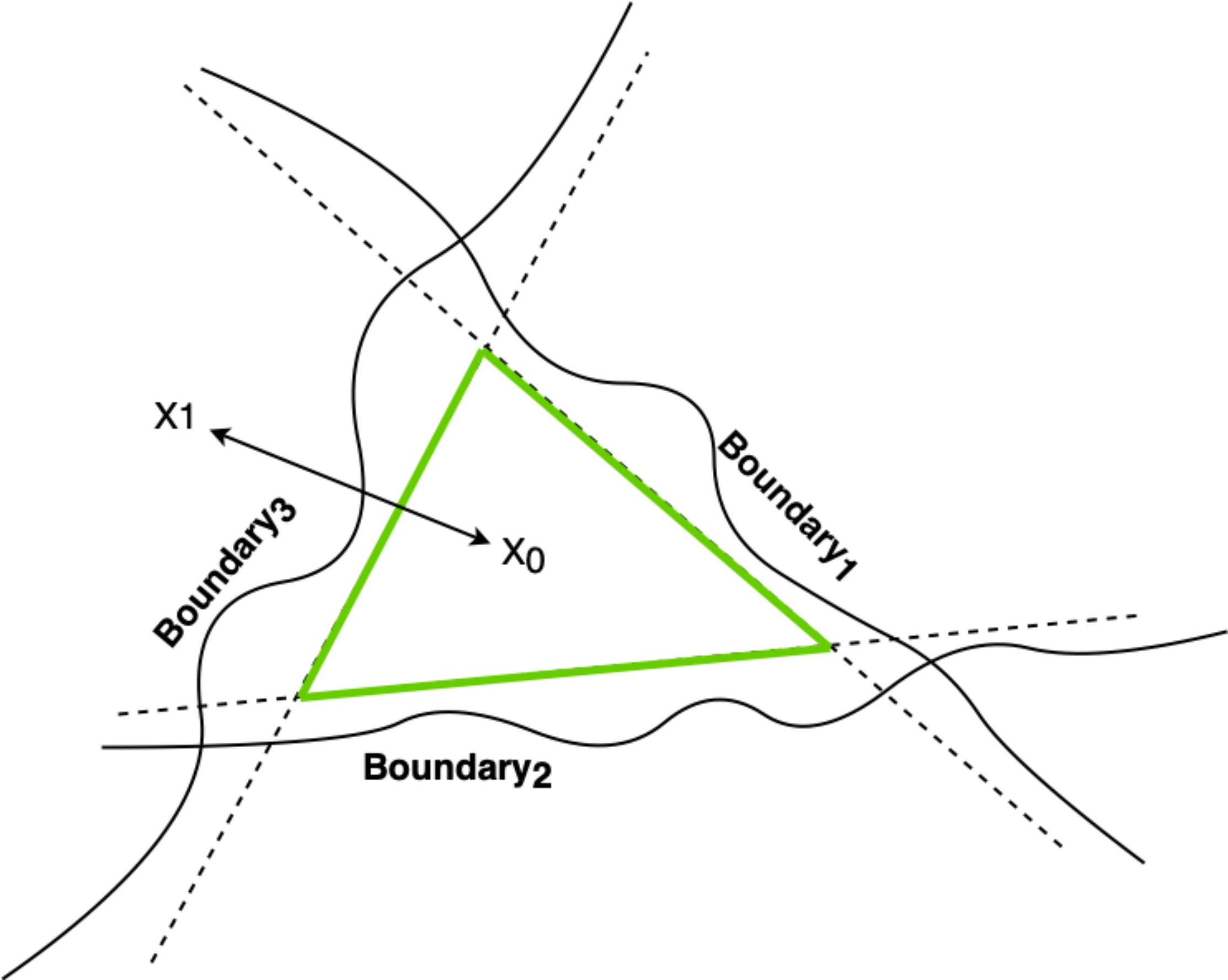}
    \caption{Illustration of Deepfool attack algorithm}
    \label{fig:decision_boundary}
\end{figure}

The Deepfool attack is formulated on the idea that neural network models act like linear classifiers with classes separated by a hyperplane. Starting with the initial input point $\mathbf{x_t}$, the algorithm determines the closest hyperplane and the smallest perturbation amount, which is the orthogonal projection to the hyperplane, at each iteration. The algorithm then computes $\mathbf{x}_{t+1}$ by adding the smallest perturbation to the $\mathbf{x}_{t}$ and checks for misclassification. The illustration of this attack algorithm is provided in Figure \ref{fig:decision_boundary}. This attack can break defensive distillation method and achieves higher success rates than previously mentioned iterative attack approaches. But the downside is, produced adversarial sample generally lies close to the decision boundary of the model.

\subsubsection{Carlini{\&}Wagner Attack}
\label{sec:carlini}

Proposed by Carlini and Wagner \cite{carlini2017evaluating}, and it is one of the strongest attack algorithms so far. As a result, it's commonly used as a benchmark for the adversarial defense research groups, which tries to develop more robust DNN architectures that can withstand adversarial attacks. It is shown that, for the most well-known datasets, the CW attack has a greater success rate than the other attack types on normally trained models. Like Deepfool, it can also deceive defensively distilled models,  which other attack types struggle to create adversarial examples for.

In order to generate more effective and strong adversarial samples under multiple $l_{p}$ norms, the authors reformulate the attack as an optimization problem which may be solved using gradient descent. A $confidence$ parameter in the algorithm can used to change the level of prediction score for the created adversarial sample. For a normally trained model, application of CW attack with default setting (confidence set to 0) would generally yield to adversarial samples close to decision boundary. And high confident adversaries generally located further away from decision boundary. 

 Adversarial machine learning is a burgeoning field of research, and we see a lot of new adversarial attack algorithms being proposed. Some of the recent remarkable ones are: i) Square Attack \cite{andriushchenko2020square} which is a query efficient black-box attack that is not based on model's gradient and can break defenses that utilize gradient masking, ii) HopSkipJumpAttack \cite{9152788} which is a decision-based attack algorithm based on an estimation of model's gradient direction and binary-search procedure for approaching the decision boundary, iii) Prior Convictions \cite{ilyas2019prior} which utilizes two kinds of gradient estimation (time and data dependent priors) and propose a bandit optimization based framework for adversarial sample generation under loss-only access black-box setting and iv) Uncertainty-Based Attack \cite{tuna2021exploiting} which utilizes both the model's loss function and quantified epistemic uncertainty to generate more powerful attacks.
 
\subsection{Adversarial defense}

\subsubsection{Defensive Distillation}

Although the idea of knowledge distillation was previously introduced by Hinton et al. \cite{hinton2015distilling} to compress a large model into a smaller one, the utilization of this technique for adversarial defense purposes was first suggested by Papernot et al. \cite{papernot2016distillation}. The algorithm starts with training a $teacher \  model$ on training data by employing a high temperature (T) value in the softmax function as in Equation \ref{eq:softmax_T}, where $p_{i}$ is the probability of i\textsuperscript{th} class and $z_{i}$'s are the logits.

\begin{equation}
 p_{i}  = \frac{\exp(\frac{z_{i}}{T})}{\sum_{j} \exp(\frac{z_{i}}{T})} 
 \label{eq:softmax_T}
\end{equation}

\begin{figure}[!htbp]
 \centering
	\includegraphics[width=1.0\linewidth]{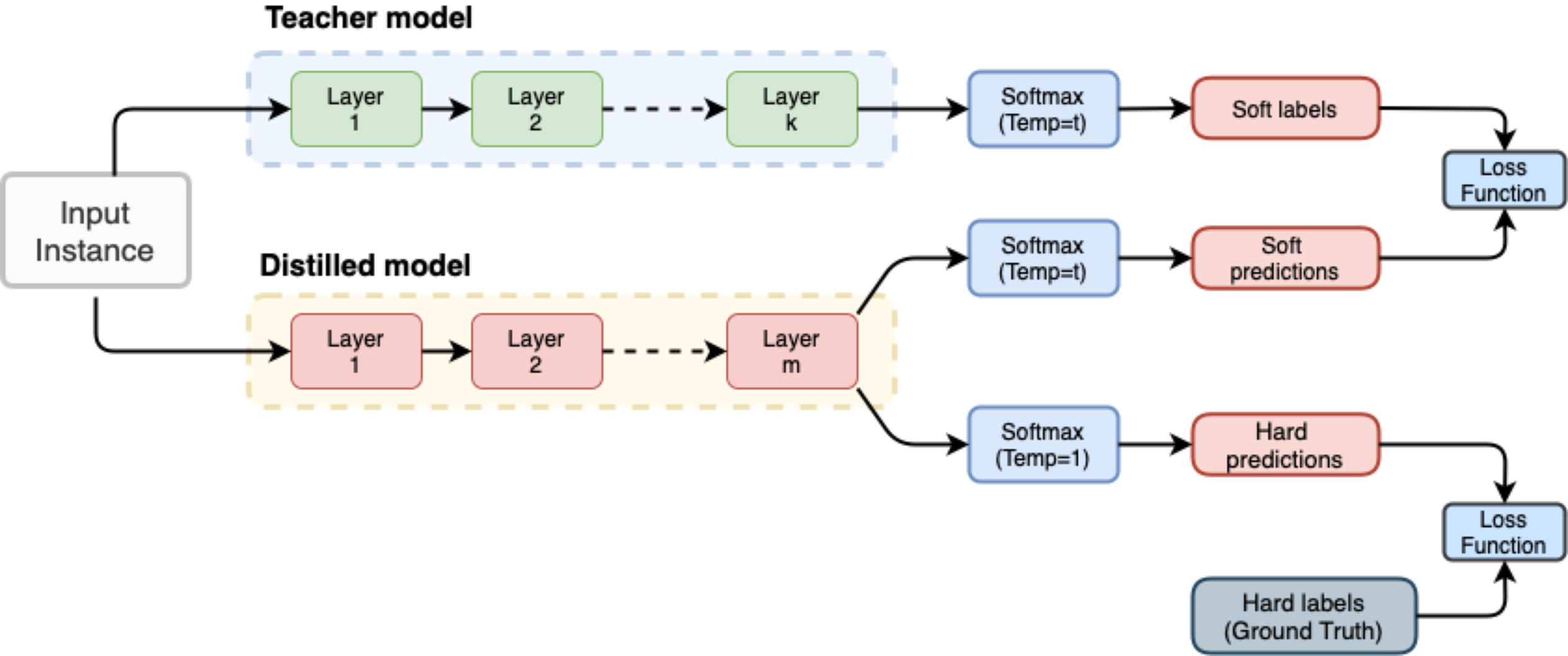}
	\caption{Defensive Distillation.}
	\label{fig:edefens-dist}
\end{figure}

Then, using the previously trained teacher model, each of the samples in the training data is labeled with soft labels calculated with temperature (T) in prediction time. The $distilled \ model$ is then trained with the soft labels acquired from the teacher model, again with a high temperature (T) value in the softmax. When the training of the student model is over,  we use temperature value as 1 during prediction time. Figure \ref{fig:edefens-dist} shows the overall steps for this technique.

\subsubsection{Adversarial Training}

Adversarial training is an intuitive defense method in which the model's robustness is increased by training it with adversarial samples. As demonstrated in Eq. \ref{eq:min_maxx}, this strategy can be mathematically expressed as a Minimax game.

\begin{equation}
\label{eq:min_maxx}
\underset{\theta}{min} \ \underset{|\delta\| \leq \epsilon }{max} \  J(h_\theta(x+\delta), y)
\end{equation}

where $h$ denotes the model, $J$ denotes the model's loss function, $\theta$ represents model's weights and y is the actual label. $\delta$ is the amount of perturbation amount added to input x and it is constrained by given $\epsilon$ value. The inner objective is maximized by using the most powerful attack possible, which is mostly approximated by various adversarial attack types. In order to reduce the loss resulting from the inner maximization step, the outside minimization objective is used to train the model. This whole process produces a model that is expected to be resistant to adversarial attacks used during the training of the model. For adversarial training, Goodfellow et al. \cite{goodfellow2015explaining} used adversarial samples crafted by the FGSM attack. And Madry et al. used the PGD attack \cite{madry2019deep} to build more robust models, but at the expense of consuming more computational resources. Despite the fact that adversarial training is often regarded as one of the most effective defenses against adversarial attacks, adversarially trained models are nevertheless vulnerable to attacks like CW.

Adversarial ML is a very active field of research, and new adversarial defense approaches are constantly being presented. Among the most notable are: i) High-Level Representation Guided Denoiser (HGD) \cite{liao2018defense} which avoids the error amplification effect of a traditional denoiser by utilizing the error in the upper layers of a DNN model as loss function and manages the training of a more efficient image denoiser, ii) APE-GAN \cite{shen2017apegan} which uses a Generative Adversarial Network (GAN) trained with adversarial samples to eliminate any adversarial perturbation of an input image, iii) Certified Defense \cite{raghunathan2020certified} which proposes a new differentiable upper bound yielding a model certificate ensuring that no attack can cause the error to exceed a specific value and iv) \cite{tuna2020closeness} which uses several uncertainty metrics for detecting adversarial samples.

\section{Approach}

\subsection{Chosen Activation Functions}

We used specific type of activation functions (sigmoid and hyperbolic tangent) whose derivatives can be expressed in terms of themselves. And the derivative of these activation functions approaches to 0 when the output of the activation functions approaches to their maximum and minimum values.

\begin{figure}[!htbp]
    \centering
    \subfloat[\centering Sigmoid\label{fig:sig}]{{\includegraphics[width=5.2cm]{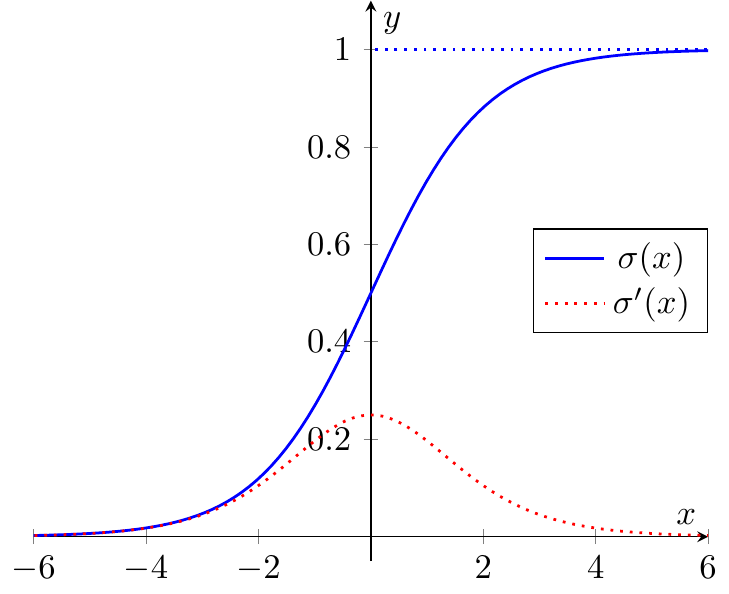} }}%
    \qquad
    \subfloat[\centering Hyperbolic Tangent\label{fig:tanh}]{{\includegraphics[width=5.2cm]{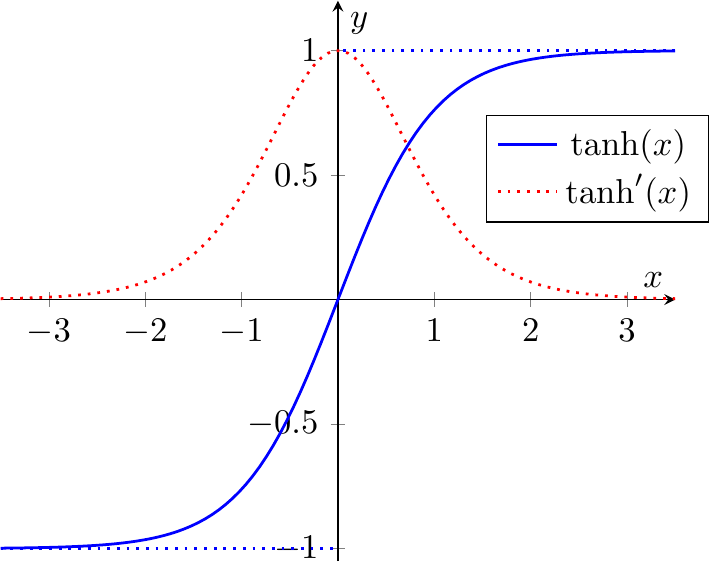} }}%
\end{figure}

Starting with the sigmoid function; we know that sigmoid function ($\sigma(x)$) can be represented as in Eq. \ref{eq:sigmoid} and it squeezes the input to the range of 0 and 1 as can be seen in Figure \ref{fig:sig}.

\begin{equation}
\sigma(x) = \frac{1}{1+e^{-x}}
\label{eq:sigmoid}
\end{equation}

And the derivative of Sigmoid function can be expressed as in Eq. \ref{eq:sigmoid_derivative}:

\begin{equation}
\frac{d}{{dx}}\sigma(x) = \sigma(x) . (1 - \sigma(x))
\label{eq:sigmoid_derivative}
\end{equation}

One can easily derive using above formulation or verify from Figure \ref{fig:sig} that the derivative of sigmoid function approach to 0 when the output of the sigmoid function approaches to 0 or 1.

Similarly, we can represent hyperbolic tangent ($\tanh(x)$) function as in Eq. \ref{eq:tanhh}. Different from sigmoid, hyperbolic tangent function squeezes the input to the range of -1 and 1 as can be seen in Figure \ref{fig:tanh}.

\begin{equation}
\tanh{x}=\frac{e^x - e^{-x}}{e^x + e^{-x}}
\label{eq:tanhh}
\end{equation}

The derivative of hyperbolic tangent function can be expressed as in Eq. \ref{eq:tanhh_derivative}. Using Eq. \ref{eq:tanhh_derivative} or Figure \ref{fig:tanh}, we can verify that the derivative of $\tanh$ function approaches to 0 when the output of the $\tanh$ function approaches to -1 or 1. So, the pattern is similar to the one we see in sigmoid function. The derivative of both of these activation functions yields to 0 when the output of the activation functions are at their minimum or maximum values. This property will be quite usefull when use these activation functions at the output layer of DNN classifiers to zeroing out the gradients.

\begin{equation}
\frac{d}{{dx}}\tanh x = 1 - \tanh ^2 x
\label{eq:tanhh_derivative}
\end{equation}

\subsection{Proposed Method}

We begin this part by introducing the loss calculation for a standard deep neural network classifier. Let $K$ denotes the number of output classes, $\mathcal{D} = \{(\mathbf{x}_i,\mathbf{y}_i)\}_{i=1}^{N}$ be our dataset, where $x_i  \in \mathbb{R}^{d}$ and $y_i  \in \{o_1,o_2...,o_k\}$ are the $i^{th}$ input and output respectively where $o_k$ is the one-hot encoded vector with the only $k^{th}$ index being one and zero for the other indices and the probability output score of any output class with index $k  \in \{0,1...,K-1\}$ is represented by $P_k$. Based on this notation, the loss value ($J$) of the classifier for any test input $x^*$ can be calculated using cross-entropy loss function as below:

\begin{equation}
\begin{split}
 J =  -\sum_{k=0}^{K-1}o_k[k] \cdot \log(P_k)
   & = -\log(P_{true})
 \end{split} 
\end{equation}

\begin{figure}[!htbp]
     \centering
     \begin{subfigure}[b]{0.8\linewidth}
         \centering
         \includegraphics[width=\linewidth]{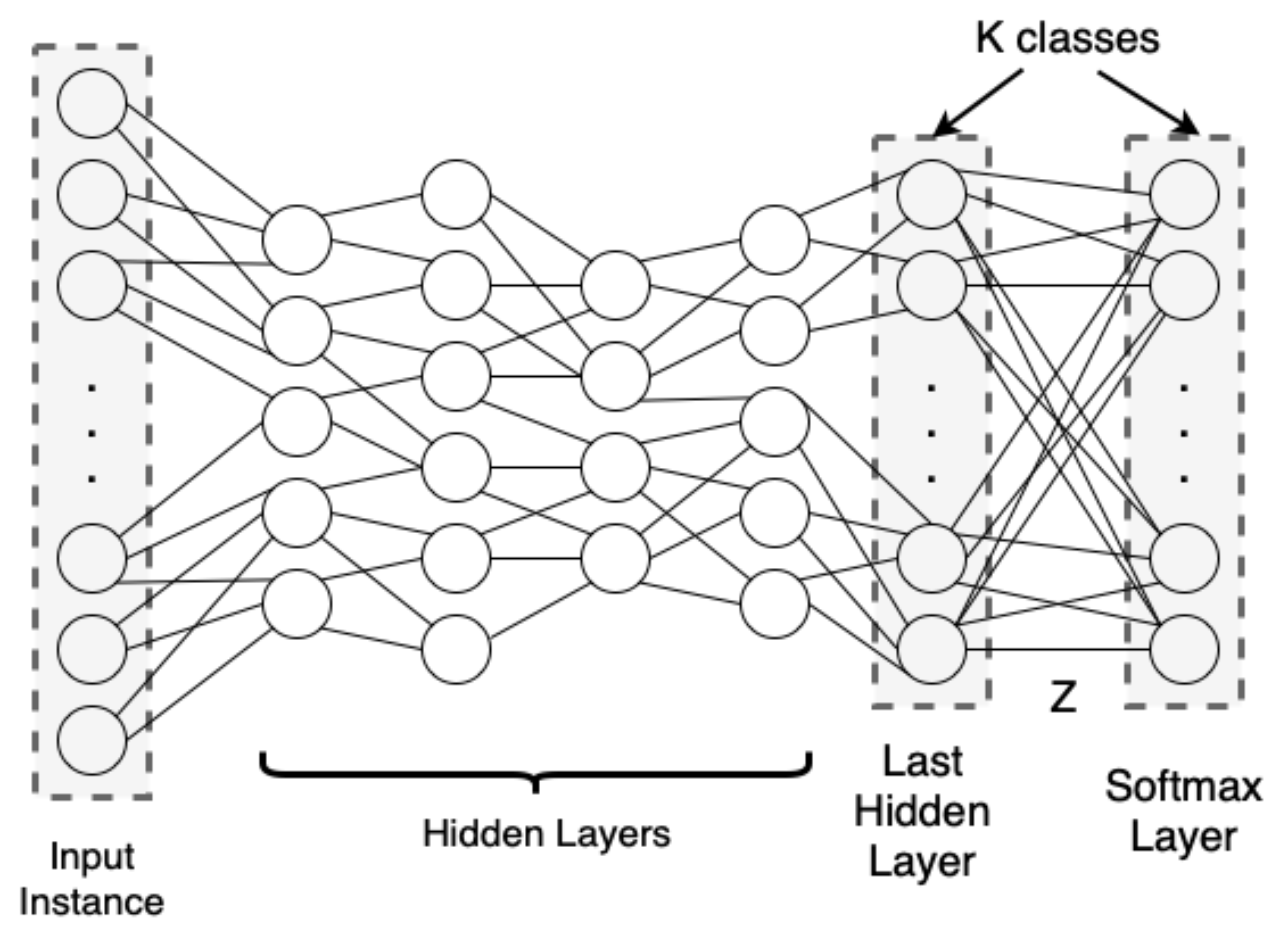}
         \caption{Standard DNN Classifier}
         \label{fig:standard_dnn}
     \end{subfigure}
     \hfill
     \begin{subfigure}[b]{0.8\linewidth}
         \centering
         \includegraphics[width=\linewidth]{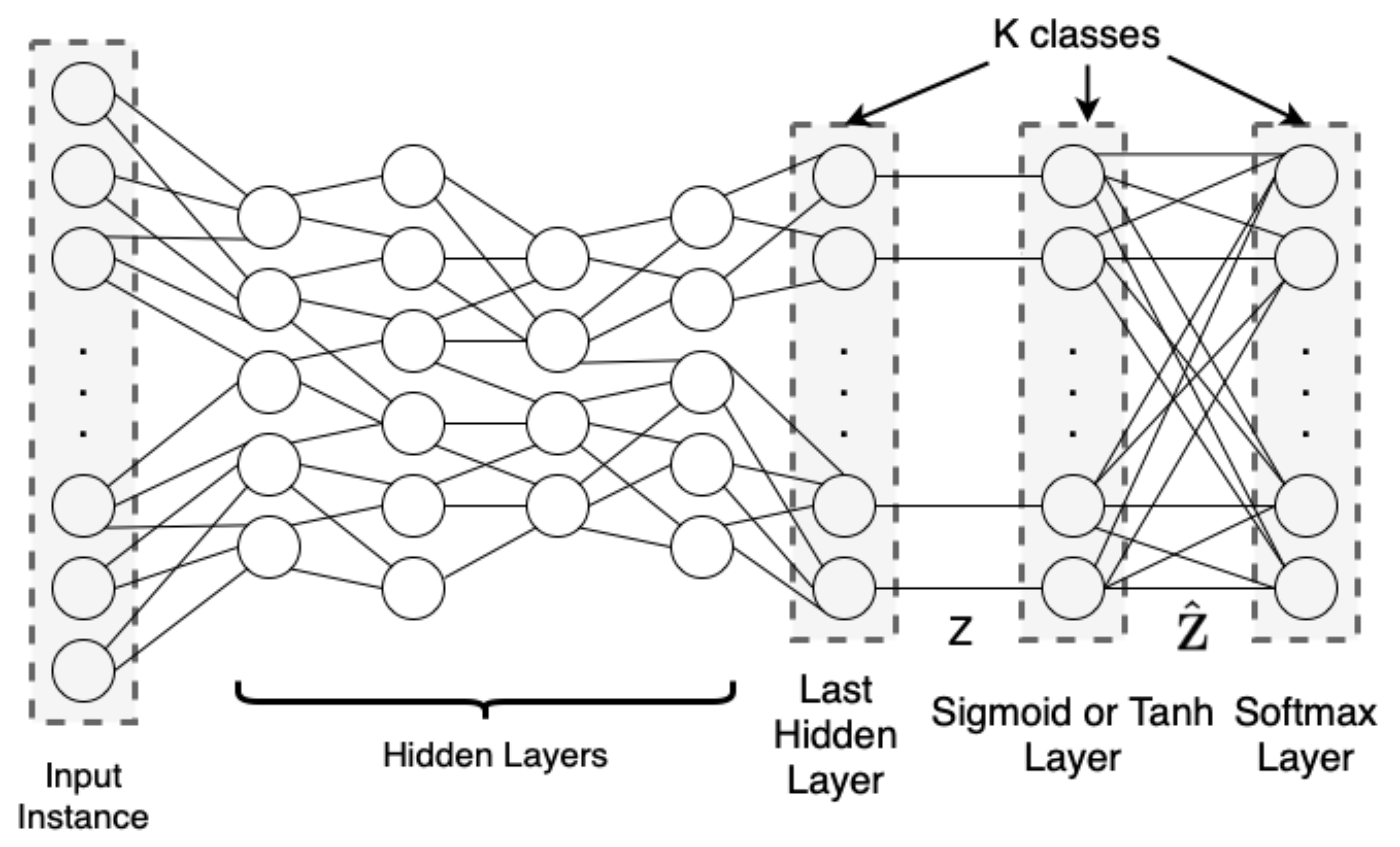}
         \caption{The proposed classifier}
         \label{fig:proposed_dnn}
     \end{subfigure}
        \caption{Comparison of standard DNN classifier and the proposed classifier}
        \label{fig:proposed_classifier}
\end{figure}

As can be seen in Figure \ref{fig:standard_dnn}, in standard DNN-based classifiers that are widely used today, usage of activation functions in the output layer is omitted and the prediction scores of each class is calculated by feeding the output of the last layer of the network (logits) to the softmax function. If we denote the logits by $Z= \{z_0,z_1...,z_{K-1}\}$, we can calculate the derivative of the loss with respect to $k^{th}$ logit using Eq. \ref{eq:logit_derivative}. Formal derivation of the Eq. \ref{eq:logit_derivative} is provided in Appendix B. 

\begin{equation}
\frac{\partial J}{\partial z_k} = P_k - o_k[k]
 \label{eq:logit_derivative} 
\end{equation}

Loss-based adversarial attacks try to exploit the gradient of the loss function $(J)$ with respect to input sample $x$,  and what the attacker is trying to do is to use $\frac{\partial J}{\partial x}$ to maximize $J$. We know from chain rule that $\frac{\partial J}{\partial x} = \frac{\partial J}{\partial z}.\frac{\partial z}{\partial x}$. Therefore, for any target class $k$, the gradient of the model's loss function with respect to the input image is directly proportional to $\frac{\partial J}{\partial z_k}$.  In response to such kind of attack idea, several defense approaches have been proposed which mask the gradients of the models. For example, Defensive Distillation technique achieves this against untargeted loss-based attacks by enabling the model to make highly confident predictions. Because, when the model makes highly confident predictions in favor of the true class; $P_{true}$ approaches 1, and since the label for true class is also 1, $\frac{\partial J}{\partial z}$ and therefore $\frac{\partial J}{\partial x}$ approaches to 0 for the specific untargeted attack case. 

However, above approach will not work for targeted attack case. Because, in order to prevent targeted attacks, we have to make $\frac{\partial J}{\partial z}$ to become 0 for target class. And, the way to achieve this for standard DNN-based classifiers is to make target probability ($P_{target}$) to be very close to 1 (to make $P_{target} - o_{target}[target]$ equals to 0) which obviously contradicts with the natural learning task. Therefore, there actually exists a dilemma between masking the gradient of the model for targeted attack case and achieving the task of learning for the model at our hand. This phenomenon is beautifully explained by Katzir et al. in \cite{katzir2019blocking}. 

To overcome this problem, we propose to use either of the two commonly known nonlinear activation functions (sigmoid and $\tanh$) on the logits of the model as depicted in Figure \ref{fig:proposed_dnn}. The important thing is to apply an high temperature value to these activation functions during learning process (e.g. : $\sigma(x,T)=1/(1+\exp(-x/T))$ and use the model by ignoring the temperature value at prediction time, just like defensive distillation technique. After our proposed modification, the output of the last layer will be $\hat{Z}$, where  $\hat{Z}=\{\hat{z}_0,\hat{z}_1...,\hat{z}_{K-1}\}$ and $\hat{Z}=\tanh{(Z)}$ or $\hat{Z}=\sigma{(Z)}$, depending on the chosen activation function. Based on this modified architecture, derivative of the model's loss with respect to input image under gradient-based attack against any class $k$ can be formulated as below:

\begin{equation}
\begin{split}
\frac{\partial J}{\partial x} = \frac{\partial J}{\partial \hat{z}_k}.\frac{\partial \hat{z}_k}{\partial z_k}.\frac{\partial z_k}{\partial x}
\end{split}
\label{eq:long_chain}
\end{equation}

In case of sigmoid function, the above equation can be reformulated as below by using Eq. \ref{eq:sigmoid_derivative} and  Eq.\ref{eq:logit_derivative}.

\begin{equation}
\begin{split}
\frac{\partial J}{\partial x} = (P_k - o_k[k]).\hat{z}_k.(1 - \hat{z}_k).\frac{\partial z_k}{\partial x}
\end{split}
\label{eq:long_sigmoid}
\end{equation}

And in case of tanh function, the Eq.\ref{eq:long_chain} can be written as below Eq. \ref{eq:tanhh_derivative} and  Eq.\ref{eq:logit_derivative}:

\begin{equation}
\begin{split}
\frac{\partial J}{\partial x} = (P_k - o_k[k]).(1 - \hat{z}_k ^2).\frac{\partial z_k}{\partial x}
\end{split}
\label{eq:long_tanh}
\end{equation}

During the training of the DNN classifier depicted in Figure \ref{fig:proposed_dnn}, we force $\hat{z_k}$ to be at it's maximum possible value for the true class in order the maximize the final softmax prediction score. And similarly, we force $\hat{z_k}$ to be at it's minimum value for the other classes. Therefore, in case of sigmoid and tanh functions, $\hat{z_k}$ will approach to 1 for true class. And for the classes other than the true class, $\hat{z_k}$ will approach to 0 and -1 for sigmoid and tanh functions respectively. Since we additionally applied a high temperature value to these activation functions during training time, the output of these activation functions ($\hat{z_k}$) will be even more close to their maximum and minimum values at prediction time when we omit their temperature values. Consequently, Eq. \ref{eq:long_sigmoid} and Eq. \ref{eq:long_tanh} will approach to 0 for both targeted and untargeted attack cases. Because, if we use the proposed model architecture using sigmoid function, $\hat{z}_k.(1 - \hat{z}_k)$ will be 0 when $\hat{z}_k$ is either 0 or 1. And if we use the proposed model architecture using tanh function, $1 - \hat{z}_k ^2$ will become 0 when $\hat{z}_k$ is either -1 or 1. This way, we can successfully zero out (mask) the gradients of the model for loss-based targeted and untargeted attack threats. To avoid any round-off errors in floating point operations, high precision should be set for floating point numbers in the ML calculations.

\subsection{Visual Representations of Loss Surfaces}

\begin{figure*}[!htbp]
    \centering

    \subfloat[\centering normal model untargeted\label{fig:normal_untargeted}]{{\includegraphics[width=5cm]{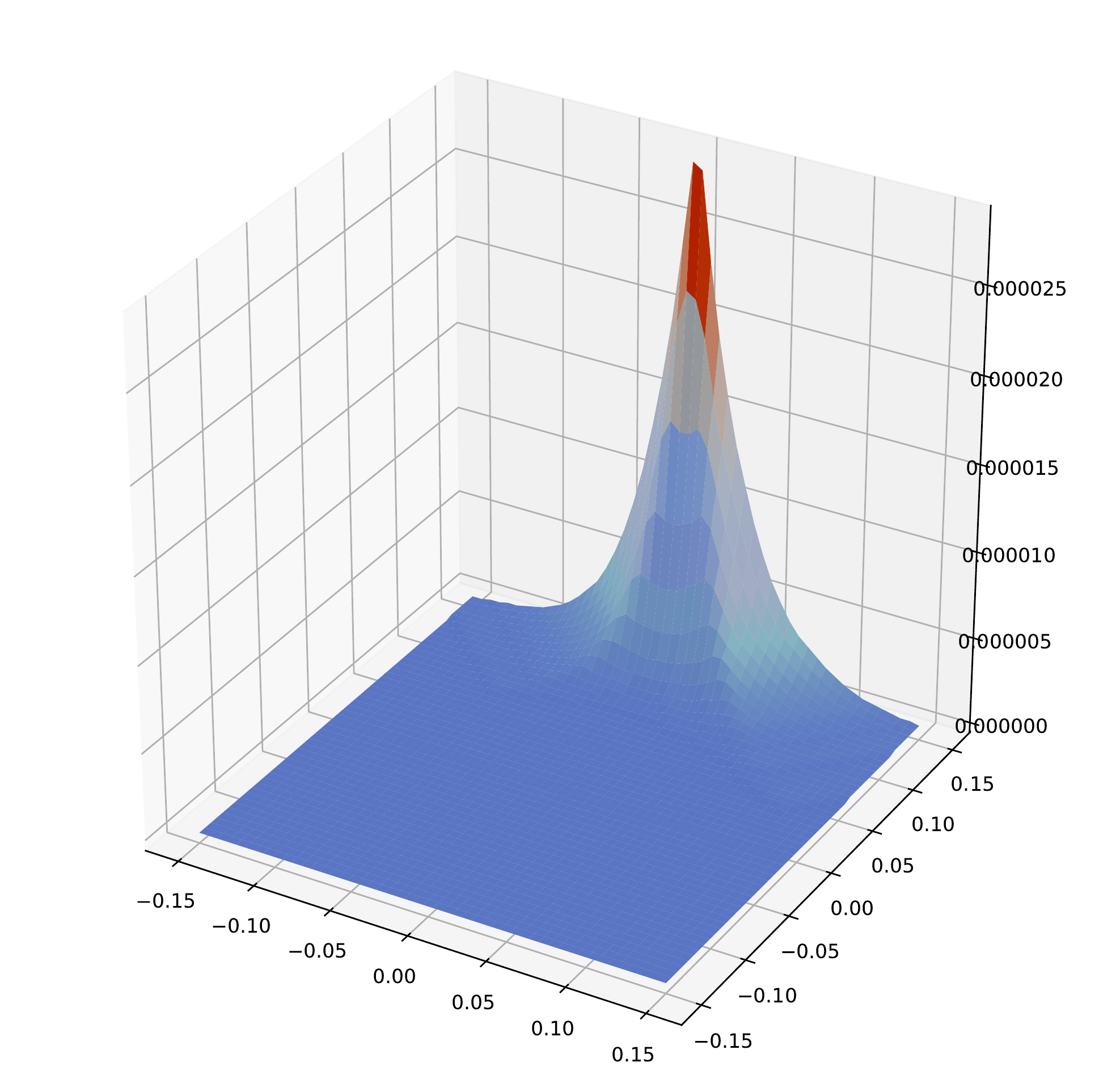} }}%
    \qquad
    \subfloat[\centering normal model targeted\label{fig:normal_targeted}]{{\includegraphics[width=5cm]{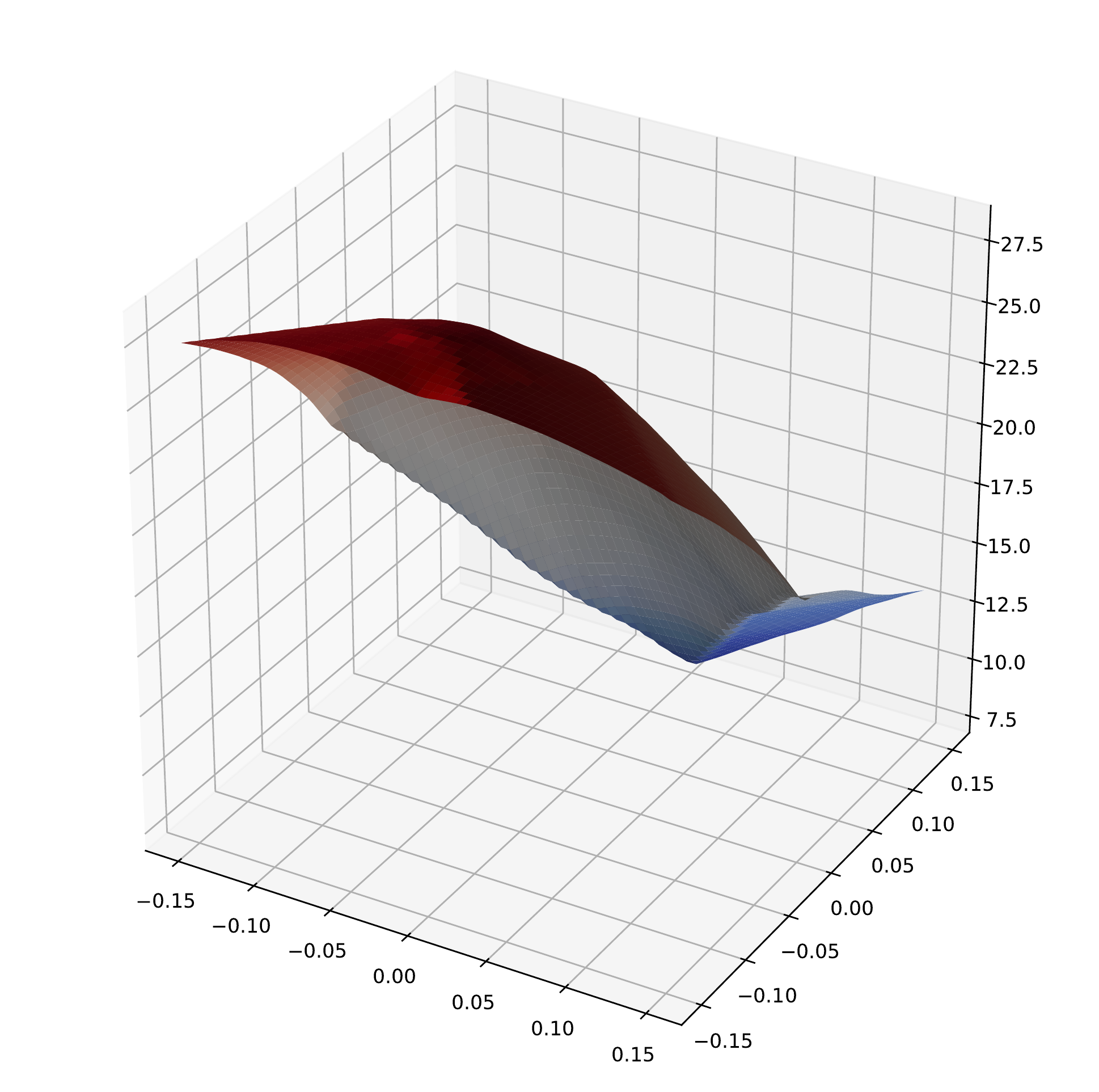} }}%

    \subfloat[\centering distilled model untargeted\label{fig:distilled_untargeted}]{{\includegraphics[width=5cm]{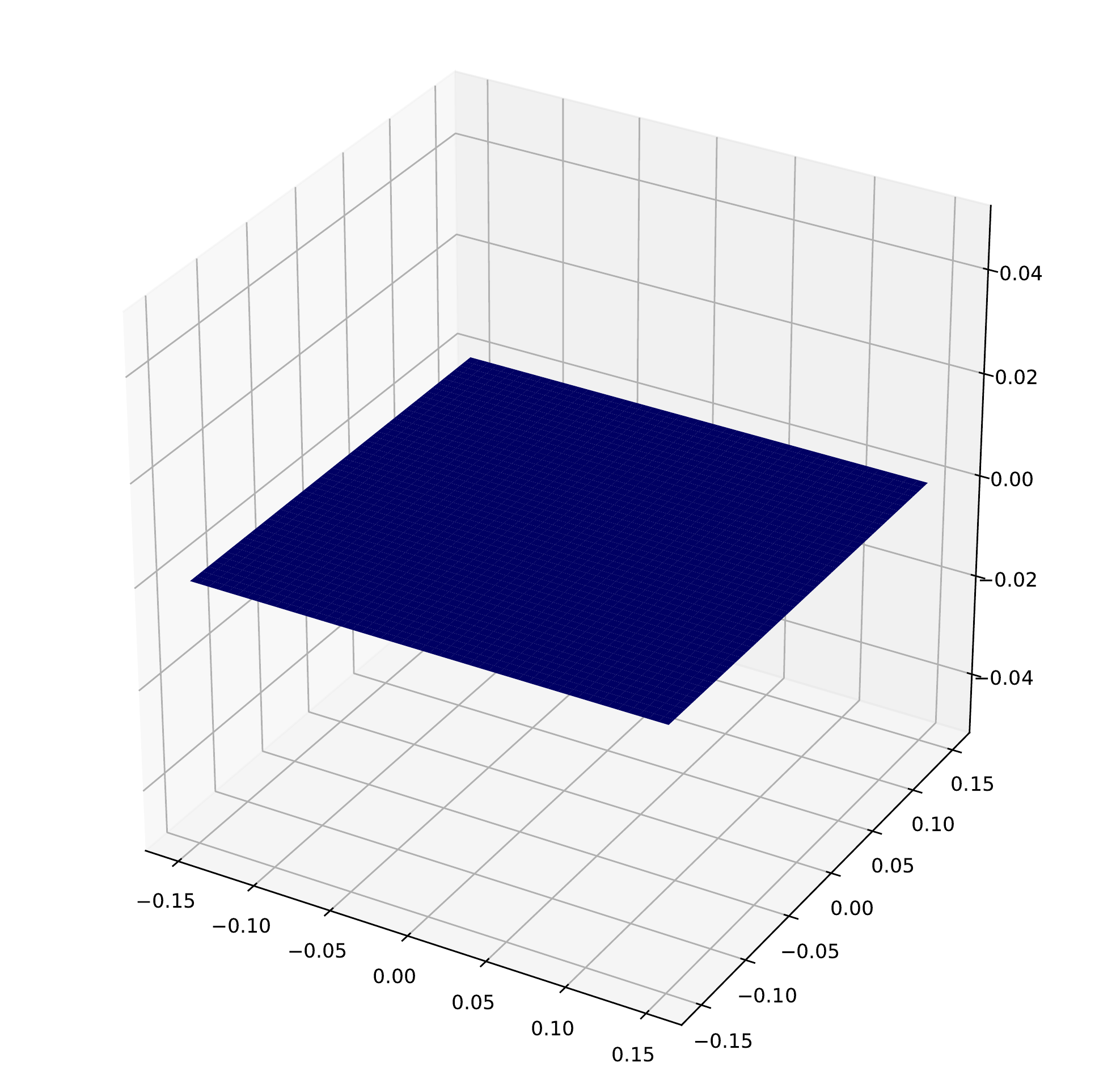} }}%
    \qquad
    \subfloat[\centering distilled model targeted\label{fig:distilled_targeted}]{{\includegraphics[width=5cm]{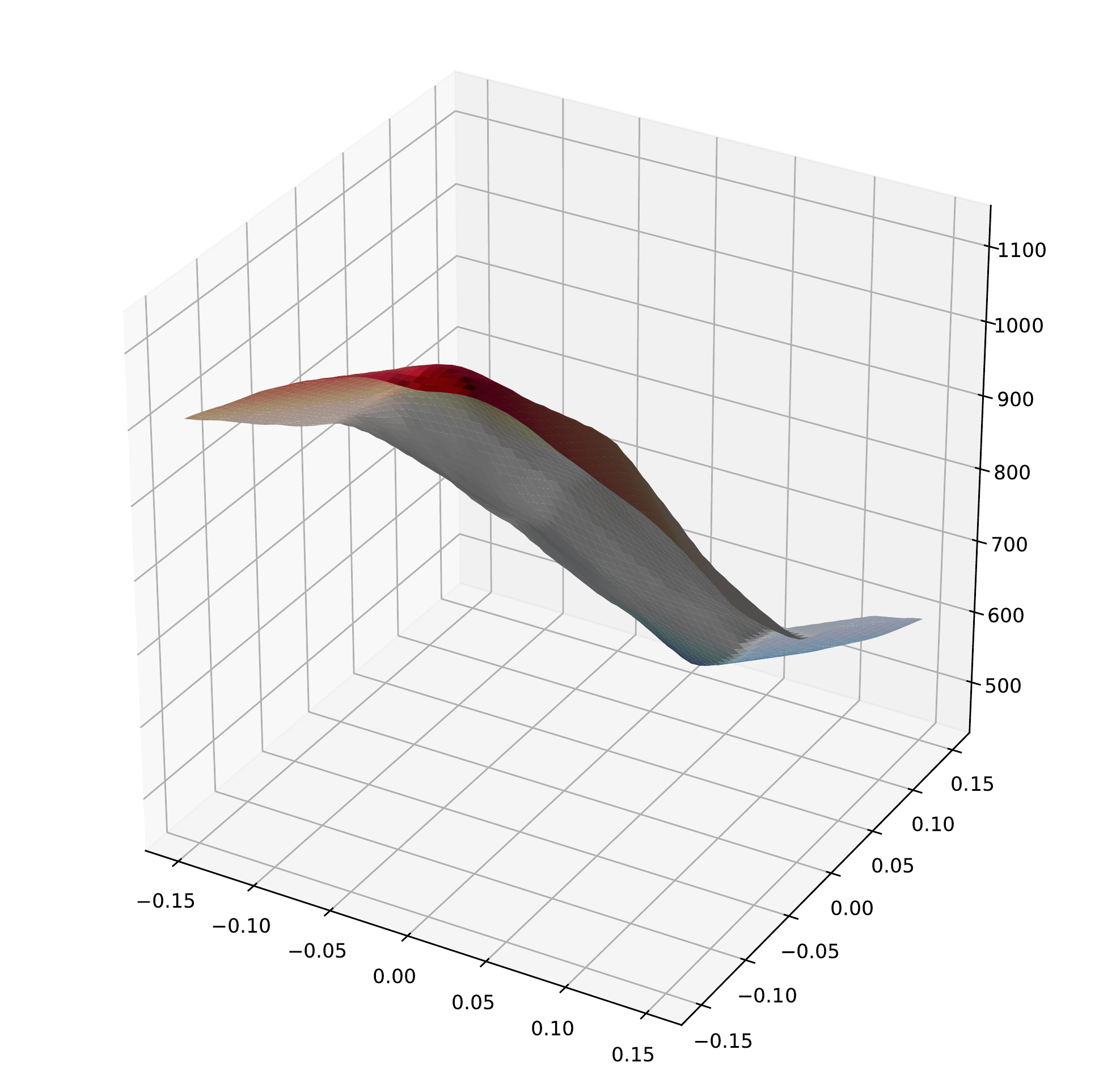} }}%

    \subfloat[\centering our model untargeted\label{fig:our_untargeted}]{{\includegraphics[width=5cm]{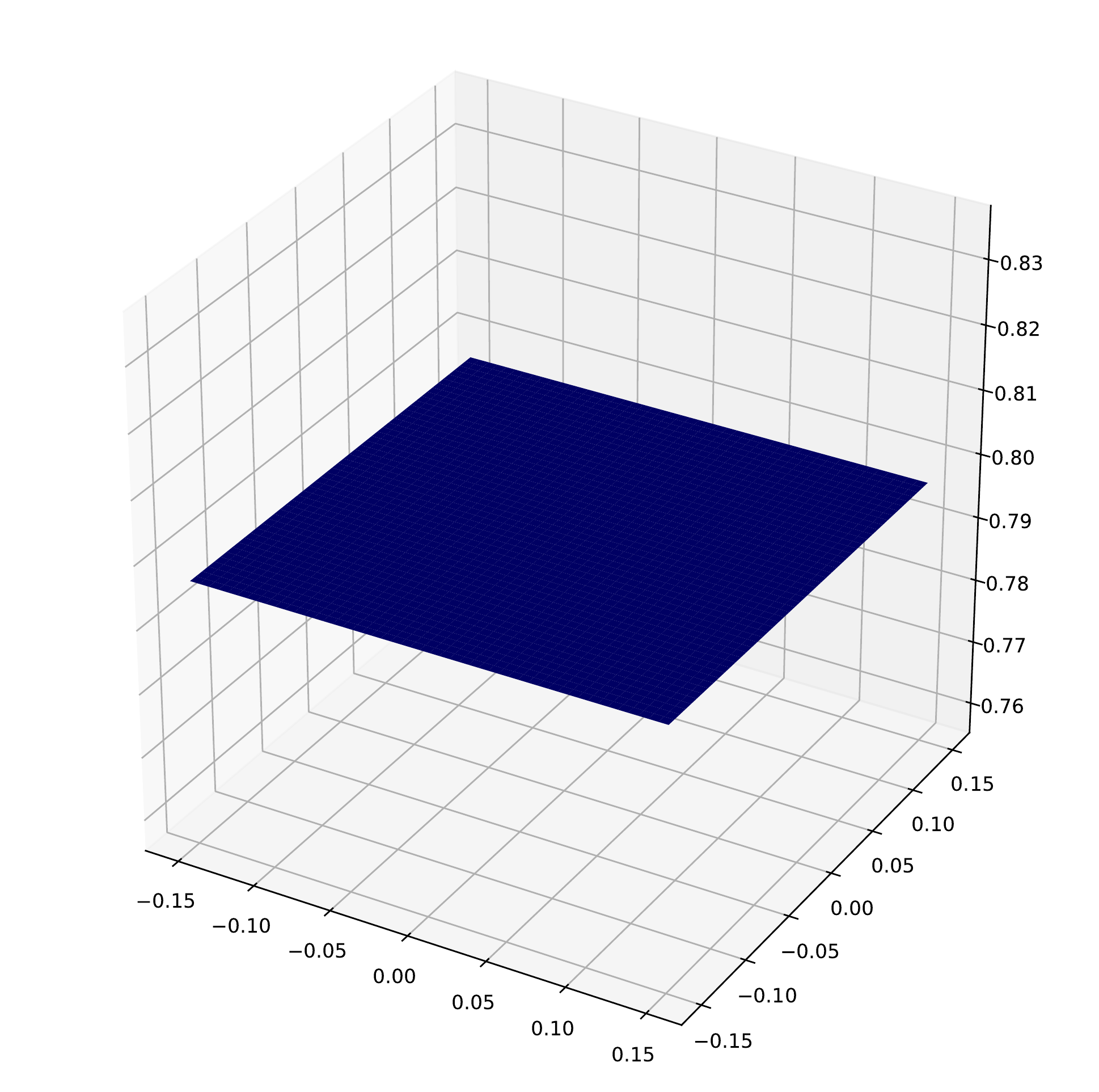} }}%
    \qquad
    \subfloat[\centering our model targeted\label{fig:our_targeted}]{{\includegraphics[width=5cm]{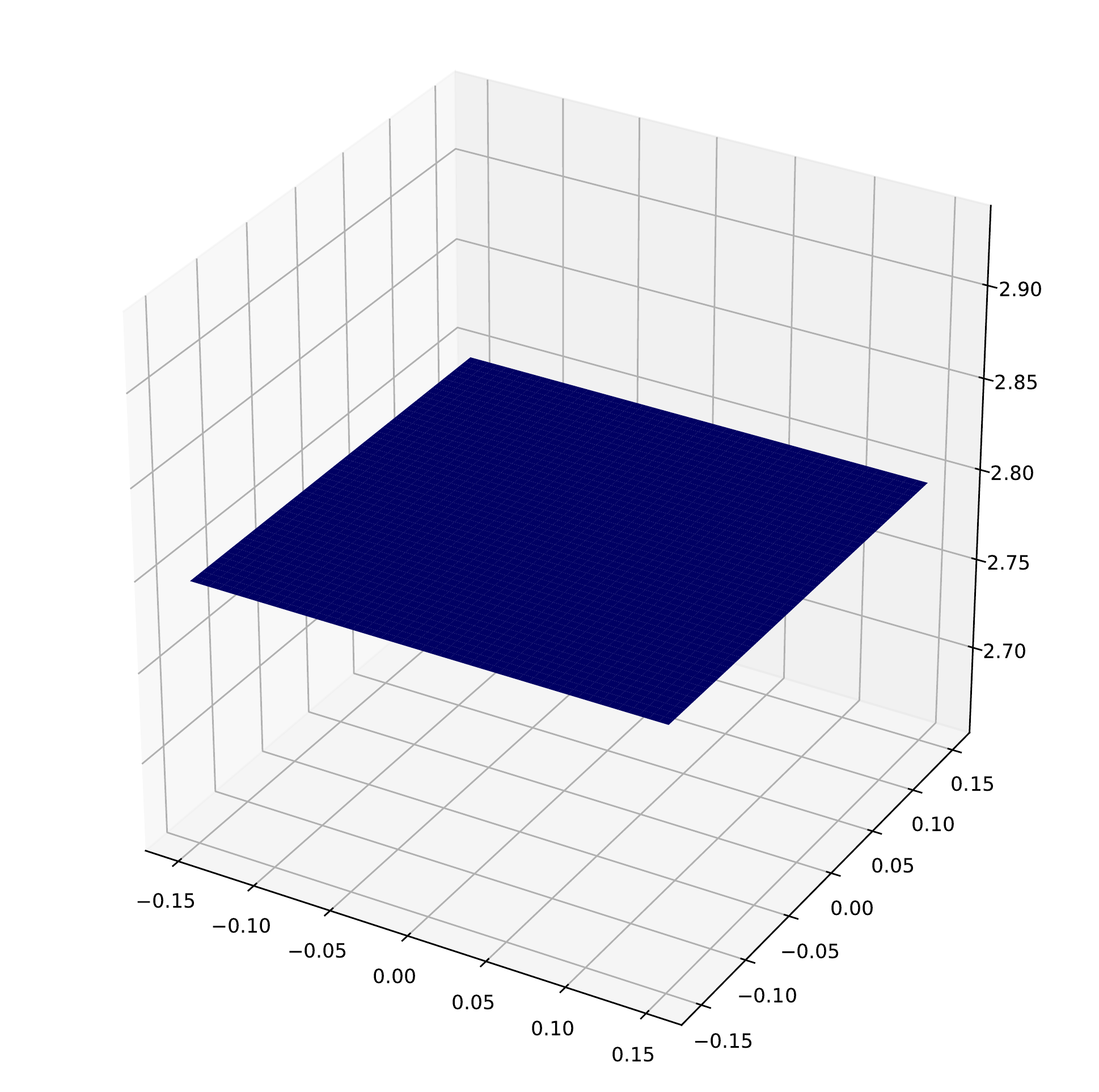} }}%
    \caption{Loss surfaces of various models under untargeted and targeted attack scenarios}%
    \label{fig:loss-surfaces}%
\end{figure*}

We know that normally-trained models are vulnerable to gradient-based white-box targeted and untargeted attack threats. The main reason for this vulnerability lies in the ability of the attacker to successfully exploit the loss function of the model. To illustrate this fact, we made a simple experiment using a test image from MNIST (Digit) dataset and draw the loss surfaces of various models against two different directions (one for loss gradient direction and one for a random direction). When we check Figures \ref{fig:normal_untargeted} and \ref{fig:normal_targeted} which display the loss surfaces of normally-trained model, we see in both cases that there exists a useful gradient information which the attacker can exploit to craft both untargeted and targeted adversarial samples.

To prevent the above vulnerability, various defense methods has been proposed, including Defensive distillation. This technique was found to significantly reduce the ability of traditional gradient-based untargeted attacks to build adversarial samples. Because defense distillation has an effect of diminishing the gradients down to zero for untargeted attack case and the usage of standard objective function is not effective anymore. As depicted in Figure \ref{fig:distilled_untargeted}, the gradient of the distilled model diminishes to zero and thus loss-based untargeted attacks have difficulty in crafting adversarial samples for defensively distilled models. However, it was then demonstrated that attacks, such as the TGSM attack, could defeat the defensive distillation strategy \cite{ross2017improving}, but without providing a mathematical proof about why these attacks actually work. And the actual reason of success for these kind of attacks against defensively distilled model is shown to lie in the targeted nature of these attacks \cite{katzir2019blocking}. Figure \ref{fig:distilled_targeted} demonstrates the loss surface of a distilled model under a targeted attack and we can easily see that the gradient of the model loss does not diminish to zero as in Figure \ref{fig:distilled_untargeted}. The result is not surprising at all, because for a defensively distilled model under targeted attack, we expect $P_{target}$ to be almost 0 and $o_{target}[target]$ is 1. Therefore, we expect $\frac{\partial J}{{\partial z}}$  (equivalent to $P_{target}$-$o_{target}[target]$) approach to -1 which is more than enough to exploit the gradient of the loss function for a successful attack.

As a last attempt, we analyze the loss surfaces of one of our proposed models (model which was trained using $\tanh$ activation function with high temperature value at the output layer). When we check Figures \ref{fig:our_untargeted} and \ref{fig:our_targeted}, we see that the gradient of model's loss function diminishes to 0 for both of the untargeted and targeted attack cases. And this prevents the attacker to exploit the gradient information of the model to craft successful adversarial perturbations.

\subsection{Softmax prediction scores of proposed architectures}

\begin{figure}[!htbp]
    \centering
    \includegraphics[width=1\linewidth]{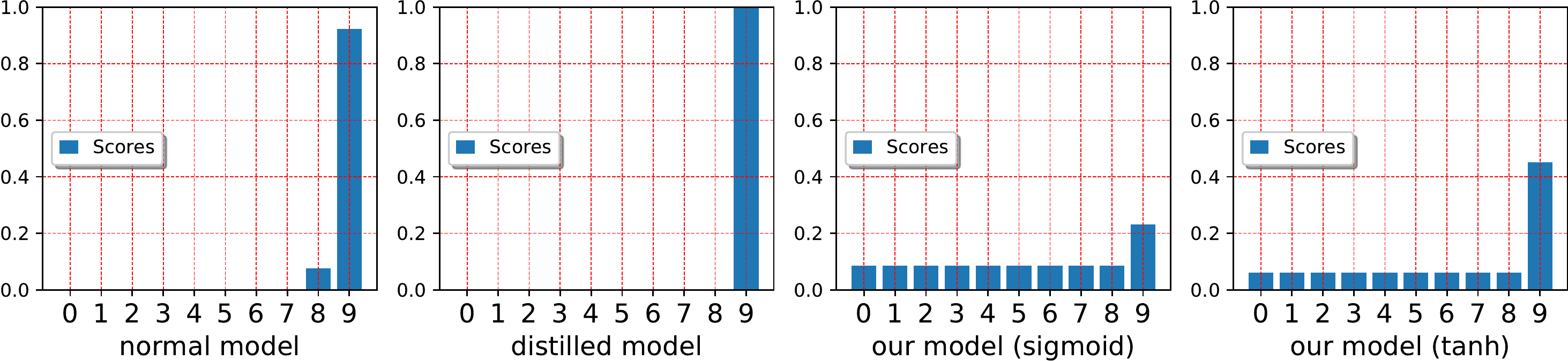}
    \caption{Softmax score outputs of various models}
    \label{fig:softmax-scores}
\end{figure}

For a normally trained standard DNN-based classifiers, we expect the model to make a prediction in favor of true class with a prediction score usually close to 1. In case of a defensively distilled model, we force the model to make high confident predictions. That is why, we see a prediction score very close to 1 in favor of the true class. However, for our proposed model architectures, the softmax prediction score of the true class is lower compared to a normal or defensively distilled model. Because the activation functions in the last layer of the model limits the values for $\hat{z}_k$ to an interval of (0,1) and (-1,1). If we use sigmoid function in the last layer, maximum prediction score will be 0.232 and if we use tanh function in the last layer, maximum prediction score will be 0.450. And this will be the case for all the predictions. Similarly, minimum prediction scores will be 0.085 and 0.061 for models with sigmoid and tanh activation functions respectively. Softmax prediction score output of a test sample from MNIST dataset is displayed in Figure \ref{fig:softmax-scores} for various models. We believe that this behaviour of our models, just like the case in defensively distilled models,  might also be quite useful to prevent attackers to infer an information that is supposed to be private from the output probability scores of any prediction of the model and might contribute to the privacy of model as suggested in \cite{shokri2017membership,shokri2019}.

\section{Experiments}

\subsection{Adversarial Assumptions}

In this research study, we assume that the attacker can chose to implement targeted or untargeted attacks towards the target the model. Our assumption was that the attacker was fully aware of the architecture and parameters of the target model as in the case of \textit{whitebox} setting and use the model as it is. Another crucial assumption concerns the constraints of the attacker. Clearly, the attacker should be limited to applying a perturbation with $l_p$ norm up to certain $\epsilon$ value for an attack to be unrecognizable to the human eye. For this study, we used $l_\infty$ and $l_2$ norm metrics to restrict the maximum perturbation amount that an adversary can apply on the input sample. Finally, the error rate of our proposed defense technique is assessed over the percentage of resulting successful attack samples which is proposed by Goodfellow et al. \cite{goodfellow2015explaining} and recommended  by Carlini et al. \cite{carlini2019evaluating}.

\subsection{Experimental Setup}\label{sec:experimental-setup}

For our experiments, we used four sets of models for each dataset as normal model, defensively distilled(student) model, proposed model with sigmoid activation and proposed model with tanh activation function at the output layer. By using same architectures, we trained our CNN models using MNIST (Digit) \cite{lecun-mnisthandwrittendigit-2010} and CIFAR-10 \cite{cifar10} datasets. For MNIST (Digit) dataset, our models attained accuracy rates of 99.35\%, 99.41\%, 98.97\% and 99.16\% respectively. And, for CIFAR-10 dataset, our models attained accuracy rates of 83.95\%, 84.68\%, 82.37\% and 80.15\% respectively. The architectures of our CNN models and the hyperparameters used in model training are listed Appendix A. Finally, we set the temperature ($T$) value as 20 and 50 for MNIST and CIFAR datasets respectively during the training of the defensively distilled model and our proposed models.

\subsection{Experimental Results}\label{sec:experimental-results}

During our tests, we only implemented attack on the test samples if our models had previously classified them accurately.  Because an adversary would have no reason to tamper with samples that have already been labeled incorrectly. For the TGSM and Targeted BIM attacks, we regard the attacks successful only if each the perturbed image is classified by the model as the chosen target class. We set the target class to "2" for MNIST (Digit) dataset, and "Cars" for CIFAR-10 dataset. We utilized an open source Python library called Foolbox \cite{rauber2018foolbox} to implement the attacks used in this study. The attack parameters used in BIM and Targeted BIM are provided in Table \ref{tab:iterative_settings}.

The results of our experiments for MNIST and CIFAR10 datasets are available in Tables \ref{tab:mnist_results},\ref{tab:mnist_results2} and Tables \ref{tab:cifar_results},\ref{tab:cifar_results2} together with the amount of perturbations applied and chosen norm metrics. Just for CW and Deepfool attacks, we used the $l_{2}$ norm equivalent of the applied perturbation by using the formula $l_{2}  = l_{inf}  \times \sqrt{n} \times \sqrt{2}/\sqrt{\pi e}$, where $n$ is the input sample dimension. When we check the results, we observe that normally trained models are vulnerable to both targeted and untargeted attack types, whereas defensively distilled models are vulnerable to only targeted attack types. And our proposed models (squeezed models) provides a high degree of robustness to both targeted (TGSM,Targeted BIM, CW) and untargeted (FGSM,BIM) attacks. This success results from the effectiveness of our models in zeroing out the gradients in both scenarios.

\begin{table}
\centering
\caption{Attack success rates on MNIST (Digit) - Part 1 }
\label{tab:mnist_results}
\scriptsize
\begin{tabular}{!{\color{black}\vrule}l|c!{\color{black}\vrule}c!{\color{black}\vrule}c!{\color{black}\vrule}c!{\color{black}\vrule}} 
\cline{2-5}
\multicolumn{1}{l!{\color{black}\vrule}}{} & \multicolumn{1}{c|}{\begin{tabular}[c]{@{}c@{}}Normal~\\Model\end{tabular}} & \begin{tabular}[c]{@{}c@{}}Distilled~\\Model\end{tabular} & \begin{tabular}[c]{@{}c@{}}Our Model~\\(Sigmoid)\end{tabular} & \begin{tabular}[c]{@{}c@{}}Our Model~\\(Tanh)\end{tabular}  \\ 
\hline
FGSM ($l_\infty$, $\epsilon$ : 0.1)                                      & 9.75\%                                                                      & 2.13\%                                                    & 0.12\%                                                        & 0.38\%                                                      \\ 
\hline
TGSM ($l_\infty$, $\epsilon$ : 0.1)                                      & 1.77\%                                                                      & 1.74\%                                                    & 0.04\%                                                        & 0.03\%                                                      \\ 
\hline
BIM  ($l_\infty$, $\epsilon$ : 0.1)                                      & 34.20\%                                                                     & 2.31\%                                                    & 0.05\%                                                        & 0.19\%                                                      \\ 
\hline
Targeted-BIM ($l_\infty$, $\epsilon$ : 0.1)                              & 13.04\%                                                                     & 9.31\%                                                    & 0.03\%                                                        & 0.02\%                                                      \\ 
\hline
CW ($l_2$, $\epsilon$ : 1.35 conf : $0$)                                                            & 80.94\%                                                                     & 59.99\%                                                   & 0.04\%                                                        & 0.07\%                                                     \\ 
\hline
Deepfool  ($l_2$, $\epsilon$ : 1.35 )                                 & 29.73\%                                                                     & 21.22\%                                                   & 0.06\%                                                        & 0.14\%                                                      \\
\hline
\end{tabular}
\label{tab:mnist_results}
\end{table}

\begin{table}
\centering
\caption{Attack success rates on MNIST (Digit) - Part 2}
\label{tab:mnist_results2}
\scriptsize
\begin{tabular}{!{\color{black}\vrule}l|c!{\color{black}\vrule}c!{\color{black}\vrule}c!{\color{black}\vrule}c!{\color{black}\vrule}} 
\cline{2-5}
\multicolumn{1}{l!{\color{black}\vrule}}{} & \multicolumn{1}{c|}{\begin{tabular}[c]{@{}c@{}}Normal~\\Model\end{tabular}} & \begin{tabular}[c]{@{}c@{}}Distilled~\\Model\end{tabular} & \begin{tabular}[c]{@{}c@{}}Our Model~\\(Sigmoid)\end{tabular} & \begin{tabular}[c]{@{}c@{}}Our Model~\\(Tanh)\end{tabular}  \\ 
\hline
FGSM ($l_\infty$, $\epsilon$ : 0.2)                                      & 31.09\%                                                                      & 2.23\%                                                    & 0.38\%                                                        & 0.12\%                                                      \\ 
\hline
TGSM ($l_\infty$, $\epsilon$ : 0.2)                                      & 9.86\%                                                                      & 8.33\%                                                    & 0.03\%                                                        & 0.04\%                                                      \\ 
\hline
BIM  ($l_\infty$, $\epsilon$ : 0.2)                                      & 98.19\%                                                                     & 2.71\%                                                    & 0.23\%                                                        & 0.08\%                                                      \\ 
\hline
Targeted-BIM ($l_\infty$, $\epsilon$ : 0.2)                              & 90.05\%                                                                     & 77.77\%                                                    & 0.03\%                                                        & 0.04\%                                                      \\ 
\hline
CW ($l_2$, $\epsilon$ : 2.70 conf : $0$)                                                            & 100\%                                                                     & 99.96\%                                                   & 0.11\%                                                        & 0.11\%                                                     \\ 
\hline
Deepfool  ($l_2$, $\epsilon$ : 2.70 )                                 & 97.69\%                                                                     & 87.41\%                                                   & 0.17\%                                                        & 0.06\%                                                      \\
\hline
\end{tabular}
\label{tab:mnist_results2}
\end{table}

\begin{table}
\centering
\caption{Attack success rates on CIFAR10 - Part 1 }
\label{tab:cifar_results}
\scriptsize
\begin{tabular}{|l|c|c|c|c|} 
\cline{2-5}
\multicolumn{1}{l|}{\begin{tabular}[c]{@{}l@{}}\\\end{tabular}} & \begin{tabular}[c]{@{}c@{}}Normal~\\Model\end{tabular} & \begin{tabular}[c]{@{}c@{}}Distilled~\\Model\end{tabular} & \begin{tabular}[c]{@{}c@{}}Our Model~\\(Sigmoid)\end{tabular} & \begin{tabular}[c]{@{}c@{}}Our Model~\\(Tanh)\end{tabular}  \\ 
\hline
FGSM ($l_\infty$, $\epsilon$ : 3/255)                                                           & ~72.38\%                                               & 13.88\%                                                   & ~4.07\%                                                       & ~1.64\%                                                     \\ 
\hline
TGSM ($l_\infty$, $\epsilon$ : 3/255)                                                           & ~22.84\%                                               & ~21.36\%                                                  & ~0.56\%                                                       & 0.23\%                                                      \\ 
\hline
BIM  ($l_\infty$, $\epsilon$ : 3/255)                                                           & ~93.53\%                                               & ~15.01\%                                                  & 2.61\%                                                        & ~0.95\%                                                     \\ 
\hline
Targeted-BIM  ($l_\infty$, $\epsilon$ : 3/255)                                                  & ~57.36\%                                               & 57.22\%                                                   & ~0.46\%                                                       & 0.22\%                                                      \\ 
\hline
CW  ($l_2$, $\epsilon$ :0.798)                                                            & ~100.00\%                                              & ~100.00\%                                                 & 2.93\%                                                        & ~1.36\%                                                     \\ 
\hline
Deepfool  ($l_2$, $\epsilon$ : 0.798)                                                      & ~99.98\%                                               & ~99.76\%                                                  & ~2.22\%                                                       & ~0.87\%                                                     \\
\hline
\end{tabular}
\end{table}

\begin{table}
\centering
\caption{Attack success rates on CIFAR10 - Part 2 }
\label{tab:cifar_results2}
\scriptsize
\begin{tabular}{|l|c|c|c|c|} 
\cline{2-5}
\multicolumn{1}{l|}{\begin{tabular}[c]{@{}l@{}}\\\end{tabular}} & \begin{tabular}[c]{@{}c@{}}Normal~\\Model\end{tabular} & \begin{tabular}[c]{@{}c@{}}Distilled~\\Model\end{tabular} & \begin{tabular}[c]{@{}c@{}}Our Model~\\(Sigmoid)\end{tabular} & \begin{tabular}[c]{@{}c@{}}Our Model~\\(Tanh)\end{tabular}  \\ 
\hline
FGSM ($l_\infty$, $\epsilon$ : 6/255)                                                           & 80.58\%                                               & 13.85\%                                                   & 4.04\%                                                       & 1.7\%                                                     \\ 
\hline
TGSM ($l_\infty$, $\epsilon$ : 6/255)                                                           & 25.41\%                                               & 25.58\%                                                  & 0.64\%                                                       & 0.3\%                                                      \\ 
\hline
BIM  ($l_\infty$, $\epsilon$ : 6/255)                                                           & 96.75\%                                               & 15.11\%                                                  & 3.07\%                                                        & 1.14\%                                                     \\ 
\hline
Targeted-BIM  ($l_\infty$, $\epsilon$ : 6/255)                                                  & 68.16\%                                               & 73.07\%                                                   & 0.6\%                                                       & 0.29\%                                                      \\ 
\hline
CW  ($l_2$, $\epsilon$ :1.596)                                                            & 100\%                                              & 100\%                                                 & 3.08\%                                                        & 1.75\%                                                     \\ 
\hline
Deepfool  ($l_2$, $\epsilon$ :1.596 )                                                      & 99.98\%                                               & 100\%                                                  & 2.17\%                                                       & 0.89\%                                                     \\
\hline
\end{tabular}
\end{table}

One other thing worth to mention about the result of our experiments is that, in addition to gradient-based attacks, our proposed models exhibit excellent performance against Deepfool attack as well. Generally, the reason behind the success of Deepfool attack against standard DNN-based classifiers is the linear nature of these models as argued by Goodfellow et. al. \cite{goodfellow2015explaining} and the authors of Deepfool paper formalized their methods based on this assumption \cite{moosavidezfooli2016deepfool}. However, since we introduce additional non-linearity to the standard DNN classifiers at the output layer, Deepfool attack algorithm fails to succeed in crafting adversarial samples compared to normally or defensively distilled models. 

\section{Conclusion}

In this study, we first showed that existing DNN-based classifiers are vulnerable to gradient-based White-box attacks. And, even the model owner uses a defensively distilled model, the attacker can still have a chance to craft successful targeted attacks. We then proposed a modification to the standard DNN-based classifiers which helps to mask the gradients of the model and prevents the attacker to exploit them to craft both targeted and untargeted adversarial samples. We empirically verified the effectiveness of our approach on standard datasets which are heavily used by adversarial ML community. Finally, we demonstrated that our proposed model variants have inherent resistance to Deepfool attack thanks to the increased non-linearity at the output layer.

In this study, we focused on securing DNN based classifiers against evasion attacks. However, it is shown that previous defense approaches on adversarial robustness suffer from privacy preservation issues\cite{shokri2019}. In the future, we plan to evaluate our proposed models against privacy related attack strategies, specifically membership inference attacks.

\section{Appendix - A}

\begin{table}[!htbp]
    \centering
    \caption{Model Architectures used in our experiments}
    \label{tab:cnn_model_arch_digit} \scriptsize
    \begin{tabular}{|c||c|c|}
        \hline
        \textbf{Dataset} & \textbf{Layer Type} &  \textbf{Layer Information}\\
        \hline \hline
        \multirow{10}{*}{MNIST - Digit} & Convolution (padding:1) + ReLU & $3 \times 3 \times 32$ \\
        & Convolution (padding:1) + ReLU & $3 \times 3 \times 32$ \\
        & Max Pooling & $2 \times 2$ \\
        & Convolution (padding:1) + ReLU & $3 \times 3 \times 64$ \\
        & Convolution (padding:1) + ReLU & $3 \times 3 \times 64$ \\
        & Max Pooling & $2 \times 2$ \\
        & Fully Connected + ReLU & $3136 \times 200$ \\
        & Dropout & p : 0.5 \\
        & Fully Connected + ReLU & $200 \times 200$ \\
        & Dropout & p : 0.5 \\
        & Fully Connected  & $200 \times 10$ \\
        \hline \hline
        \multirow{14}{*}{CIFAR10} & Convolution (Padding = 1) + ReLU & $3 \times 3 \times 32$ \\
        & Convolution (Padding = 1) + ReLU & $3 \times 3 \times 64$ \\
        & Max Pooling (Stride 2) & $2 \times 2$ \\
        & Convolution (Padding = 1) + ReLU & $3 \times 3 \times 128$ \\
        & Convolution (Padding = 1) + ReLU & $3 \times 3 \times 128$ \\
        & Max Pooling (Stride 2) & $2 \times 2$ \\
        & Convolution (Padding = 1) + ReLU & $3 \times 3 \times 256$ \\
        & Convolution (Padding = 1) + ReLU & $3 \times 3 \times 256$ \\
        & Dropout & p : 0.5 \\   
        & Max Pooling (Stride 2) & $2 \times 2$ \\        
        & Fully Connected + ReLU & $4096 \times 1024$ \\
        & Dropout & p : 0.5 \\
        & Fully Connected + ReLU & $1024 \times 256$ \\
        & Dropout & p : 0.5 \\
        & Fully Connected  & $256 \times 10$ \\
        \hline
    \end{tabular}
\end{table}

Note: The common softmax layers are omitted for simplicity. For our proposed methods, we have applied Sigmoid and Tanh activation layers just after the final fully connected layers. The model architectures are available in the shared Github repository.

\begin{table*}[!htbp]
\centering
\caption{CNN model parameters}
\label{tab:cnn_model_params}
\begin{tabular}{|l|c|c|c|c|c|c|c|c|} 
\hline
\multicolumn{1}{|c|}{\multirow{2}{*}{\begin{tabular}[c]{@{}c@{}}\\\end{tabular}}} & \multicolumn{4}{c|}{MNIST (Digit)}                                                                                                                                                                                                            & \multicolumn{4}{c|}{CIFAR-10}                                                                                                                                                                                                                \\ 
\cline{2-9}
\multicolumn{1}{|c|}{}                                                                      & \begin{tabular}[c]{@{}c@{}}Normal~\\Model\end{tabular} & \begin{tabular}[c]{@{}c@{}}Distilled~\\Model\end{tabular} & \begin{tabular}[c]{@{}c@{}}Ours\\(Sigmoid)\end{tabular} & \begin{tabular}[c]{@{}c@{}}Ours\\(Tanh)\end{tabular} & \begin{tabular}[c]{@{}c@{}}Normal\\Model\end{tabular} & \begin{tabular}[c]{@{}c@{}}Distilled\\Model\end{tabular} & \begin{tabular}[c]{@{}c@{}}Ours\\(Sigmoid)\end{tabular} & \begin{tabular}[c]{@{}c@{}}Ours \\(Tanh)\end{tabular}  \\ 
\hline
Opt.                                                                                    & Adam                                                   & Adam                                                      & Adam                                                         & Adam                                                      & Adam                                                  & Adam                                                     & Adam                                                         & Adam                                                       \\ 
\hline
LR                                                                                & 0.001                                                  & 0.001                                                     & 0.001                                                        & 0.001                                                     & 0.001                                                 & 0.001                                                    & 0.001                                                        & 0.001                                                      \\ 
\hline
Batch S.                                                                                   & 128                                                    & 128                                                       & 128                                                          & 128                                                       & 128                                                   & 128                                                      & 128                                                          & 128                                                        \\ 
\hline
Dropout                                                                                & 0.5                                                    & 0.5                                                       & 0.5                                                          & 0.5                                                       & 0.5                                                   & 0.5                                                      & 0.05                                                          & 0.25                                                        \\ 
\hline
Epochs                                                                                 & 20                                                     & 20                                                        & 20                                                           & 20                                                        & 50                                                    & 50                                                       & 50                                                           & 50                                                         \\ 
\hline
Temp.                                                                                  & 1                                                      & 20                                                        & 20                                                           & 20                                                        & 1                                                     & 50                                                       & 50                                                           & 50                                                         \\
\hline
\end{tabular}
\end{table*}

\begin{table}[!htbp]
    \centering
    \caption{Parameters that are used in BIM and Targeted BIM attacks: $\alpha$ denotes the step size and $i$ denotes \# of steps for a perturbation budget $\epsilon$}
    \label{tab:iterative_settings}
    \begin{tabular}{c|c|c} 
        \hline
        \textbf{Dataset} & \textbf{Parameters} & \textbf{$l_p$ norm}\\
        \hline \hline
        MNIST Digit & $\epsilon$ = 0.1 \& 0.2, $\alpha$ = $\epsilon$ $\cdot$ 0.1, i = 20  & $l_\infty$\\
        CIFAR10 & $\epsilon$ = 3/255 \& 6/255, $\alpha$ = $\epsilon$ $\cdot$ 0.1, i = 20 & $l_\infty$ \\
        \hline
    \end{tabular}
\end{table}

\section{Appendix - B}

The gradient derivation of the cross-entropy loss coupled with the softmax activation function is described in this part. This derivation was detailed for the first time in \cite{Campbell_onthe}. We will be using the derivation explained by Katzir et al. in \cite{katzir2019blocking} as it is. 

Softmax Function Gradient Derivation:

Let $K$ represents number of classes in training data, $y=(y_0,y_1,....,y_{K-1})$ denotes the one-hot encoded label information, $z_i$ denotes the $i^{th}$ component of the logits layer output given some network input $x$. The probability estimate of the $i^{th}$ class associated with the input by the softmax function is:

\begin{equation}
    P_i = \frac{e^{z_i}}{\sum_{k=0}^{K-1}e^{z_k}}
\end{equation}

$P_i$'s derivative with respect to $z_k$ can then be calculated as below:

\begin{equation}
    \frac{\partial P_i}{\partial z_j} = \frac{\partial \left(\frac{e^{z_i}}{\sum_{k=0}^{K-1}e^{z_k}} \right)}{\partial z_j}
\end{equation}

In the case of $i=j$, we get:

\begin{equation}
\begin{split}
    \frac{\partial P_i}{\partial z_j} & = \frac{\partial \left(\frac{e^{z_i}}{\sum_{k=0}^{K-1}e^{z_k}} \right)}{\partial z_j} = \frac{e^{z_i}\sum_{k=0}^{K-1}e^{z_k}-e^{z_i}e^{z_j}}{\left(\sum_{k=0}^{K-1} e^{z_k} \right)^2} \\
    & = \frac{e^{z_i}}{\sum_{k=0}^{K-1}e^{z_k}} \cdot \frac{(\sum_{k=0}^{K-1}e^{z_k}) - e^{z_j}}{\sum_{k=0}^{K-1}e^{z_k}} = P_i(1-p_j)
\end{split}
\end{equation}

Likewise, when $i \neq j$, we get:

\begin{equation}
\begin{split}
    \frac{\partial P_i}{\partial z_j} & = \frac{\partial(\frac{e^{z_i}}{\sum_{k=0}^{K-1}e^{z_k}})}{\partial z_j} = \frac{0-e^{z_i}e^{z_j}}{(\sum_{k=0}^{K-1}e^{z_k})^2} \\
    & = \frac{-e^{z_i}}{\sum_{k=0}^{K-1}e^{z_k}} \cdot \frac{e^{z_j}}{\sum_{k=0}^{K-1}e^{z_k}} = -P_iP_j
\end{split}
\end{equation}

When we combine the two previous results, we get:

\begin{equation}
\label{eq:different}
\begin{split}
    \frac{\partial P_i}{\partial z_j} & = \begin{cases}
    P_i(1-P_j), & \text{if $i=j$}\\
    -P_iP_j, & i \neq j
  \end{cases}
\end{split}
\end{equation}

The cross-entropy loss $L$ for any input $x$ is formulated as :

\begin{equation}
    L = -\sum_{i=0}^{K-1}y_i \cdot \log(P_i)
\end{equation}

Assuming ‘log’ as natural logarithm (ln) for simplicity, we may  formulate the gradient of the cross-entropy loss with respect to the $i^{th}$ logit as below:

\begin{equation}
\begin{split}
    \frac{\partial P_i}{\partial z_j} & = \frac{\partial (-\sum_{i=0}^{K-1}y_i \cdot \log(P_i))}{\partial z_i} \\
    & = -\sum_{k=0}^{K-1}y_k \cdot \frac{\partial \log(P_k)}{\partial z_i} = -\sum_{k=0}^{K-1}y_k \cdot \frac{\partial \log(P_k)}{\partial P_k} \cdot \frac{\partial P_k}{\partial z_i} \\
    & = -\sum_{k=0}^{K-1}y_k \cdot \frac{1}{P_k} \cdot \frac{\partial P_k}{\partial z_i}
\end{split}
\end{equation}

Combining Cross-Entropy and Softmax Function Derivatives:

Knowing from Eq. \ref{eq:different} that the softmax derivative equation for the case when $i = j$ differs from the other cases, we rearrange the loss derivative equation slightly to differentiate this case from the others:

\begin{equation}
  \begin{split}
   \frac{\partial L}{\partial z_i} & = -\sum_{k=0}^{K-1}y_k \cdot \frac{1}{P_k} \cdot \frac{\partial P_k}{\partial z_i} \\
   & = -y_i \cdot \frac{1}{P_i} \cdot \frac{\partial P_i}{\partial z_i}-\sum_{k \neq i}y_k \cdot \frac{1}{P_k} \cdot \frac{\partial P_k}{\partial z_i}
\end{split} 
\end{equation}

We can now apply the derivative of softmax we derived before to obtain:

\begin{equation}
\begin{split}
    -y_i \cdot \frac{P_i(1-P_i)}{p_i} & - \sum_{k \neq i}\frac{y_k \cdot (-P_kP_i)}{P_k} = -y_i + y_iP_i + \sum_{k \neq i} y_kP_i \\
    & = P_i\left(y_i+\sum_{k \neq i}y_k\right)-y_i
\end{split}
\end{equation}

Luckily, because $Y$ is the one-hot encoded actual label vector, we know that:

\begin{equation}
    y_i + \sum_{k \neq i} y_k = \sum_{k=0}^{K-1}y_k=1
\end{equation}

and therefore we finally end up with below expression as the derivative of the loss with respect to any logit:

\begin{equation}
    \frac{\partial L}{\partial z_i} = P_i \left( y_i + \sum_{k \neq i}Y_k \right) -y_i = P_i - y_i
\end{equation}

\bibliographystyle{IEEEtran}

\bibliography{references}

\end{document}